# Dual-Part Multi-Lateral Branched Network for Multi-Class Segmentation in Cardiovascular Catheterization Angiograms

Olatunji Omisore [1,*], Ahmed Elazab [2], Ali Shahidinejad [1], Fariza Sabrina [1]

[1] Center for Machine Learning Networking and Education Technology, CQUniversity, Rockhampton, 4701, Australia.
[2] Tsinghua Shenzhen International Graduate School, Tsinghua University, Shenzhen, 518055, China
*Corresponding Author: o.omisore@cqu.edu.au

## Abstract

*Catheterisation image processing requires segmentation models that are fast, accurate and explainable. While most of the existing studies usually focus on binary segmentation, there is a recent demand for simultaneous segmentation of multiple structures found in catheterization scenes. In this study, a dual-part MLBNet architecture is designed with multi-lateral encoder blocks and multi-head decoder branches for class-aware segmentation in cardiovascular catheterization scenes. Lateral branches in the encoder enables repeated feature extraction to learn diverse shared representations, while multiple decoder heads are used to introduce class-skewed branches that specialize in different structural properties in catheterization scenes. To analyze the performances of the dual-part MLBNet architecture, several multi-class segmentation angiogram data obtained during cardiovascular catheterization in phantom models, synthetic human-simulated aorta, and animal model are used for model training and evaluation. Results obtained showed the dual-part models could effectively separate guidewire, catheter, vessels and background pixels to their classes of memberships with high probability. The results demonstrate that all models were able to distinguish the dominant background class from foreground structures with high overall accuracy.*

## 1. Introduction

Coronary artery disease, characterized by myocardial artery blockage, causes highest number of deaths amongst the cardiovascular diseases (CVDs) [1]. Traditional open-heart diagnosis and treatment require large incisions in patients' cardiac region, causing patients to face numerous challenges. Alternatively, angiography, a less invasive procedure, that involves routine catheterization of patient's vessels for diagnosis and treatment of vascular and cardiac conditions. Hence, patients experience reduced trauma and recovery time [2]. During angiography, X-ray imaging supports clinicians for intraoperative scene visualization, however they may be exposed to radiation depending on the imaging quality and their experience in catheterization scene visualization and analytics during procedures [3].

Image segmentation techniques have been developed for processing angiographic images obtained during cardiac catheterization [4]. Cardiovascular image segmentation is challenging as angiography-based images are characterized by very low contrast, noise, motion artifacts, class overlaps in cluttered background, and severe class imbalance. The complexity of the tool and vessel structures combined with the imaging technology context pose additional challenges. Thus, fast and accurate delineation of the endovascular tool and vascular structures in interventional scenes remains a critical issue. A broad literature study on segmentation and tracking of blood vessels and endovascular tools in angiographic images has been published [5-7]. Domain overview shows segmentation of the different structures are done separately, and limited attention is found on unified multi-class segmentation of vessel, endovascular tools, and background for catheterisation scene parsing. Resolving this problem could enhance downstream tasks such as tool-vessel safety analytics in interventional cardiology.

Use of classical segmentation approaches for automating visualization during catheterization focus on tracking the shape and motion of structures in angiograms. For instance, pixel-value thresholding, intensity-level analytics, or region clustering are commonly used for grayscale or RGB-based segmentation in cardiac X-ray angiograms [8]. An in-depth review of segmentation methods in this domain shows image sequences are pre-processed, and multi-scale filters are applied to segment vascular or tool structures in the angiograms [5]. To improve segmentation accuracy, some studies added image filters for preserving edges in the angiograms. Classical methods have been developed for binary class segmentation in cardiac angiograms [9]. While high segmentation performances were reported, classical techniques are commonly sensitive to blob-like non-target artifacts in the angiograms, and their performances are not generalizable [5-7, 9]. In some studies, target regions were manually defined to reduce the computational complexities involved [10]; however, this is not suitable when processing long or moving structures in catheterization angiograms.

Deep learning (DL) techniques have been developed for segmentation of endovascular tools and vascular anatomies in catheterisation angiograms [11]. For instance, Wu *et al.* [12] addressed automatic guidewire-tip segmentation using convolutional neural networks (CNN) in 2D angiograms. Gherardini *et al.* improved catheter segmentation using synthetic data and transfer learning with lightweight U-Nets, highlighting the importance of data efficiency [13]. These studies provide strong approaches for cardiovascular image segmentation; however, they primarily focus on binary-class segmentation and single-tool extraction, typically involving guidewire or catheter. CNN-based DL architectures have provided state-of-the-art segmentation results when trained with fully supervised techniques. For model training, clinical experts are required to undergo manual largescale ground-truth annotation [10].

Segmentation of multiple endovascular tools, such as catheters and guidewires, in catheterization angiograms have also been attempted in previous studies [11-14]. While these studies showed the feasibility of automatic tool visualization through image segmentation, scope of the DL techniques were largely validated for binary segmentation of a single endovascular tool, typically either guidewire or catheter, and not simultaneous multi-class segmentation. Attention-based DL architectures have been explored to improve endovascular tool segmentation, visualisation, and tracking in fluoroscopic scenes. For instance, Zhou *et al.* [15] proposed an attention recurrent network for guidewire segmentation and tracking in intraoperative fluoroscopic videos, while Wang *et al.* [16] introduced a lightweight attention-based network for segmentation and localisation of tools during procedures. These methods incorporated attention mechanisms, recurrent modelling, and lightweight feature-fusion modules for tool delineation in low-contrast and class-imbalanced imaging conditions.

The above studies focus either on background–vessel or background–tool binary segmentation. Thus, simultaneous segmentation of multiple structures in catheterization scenes remains underexplored. For multi-structure segmentation, Vlontzos *et al.* [17] extended binary class segmentation for multi-class learning. A UNet-based model was developed to produce background–catheter binary masks irrespective of the object categories in angiogram scenes. Meanwhile, catheter structure in the first frame gets transferred to later frames to delineate the blood vessel. Recently, Robertshaw *et al.* [18] implemented various DL models for three-class segmentation to aid multi-tool tip tracking in fluoroscopic images but without blood vessel segmentation. Bian [19] presented a multi-task learning model to segment vessels, catheters, and guidewire masks in fluoroscopic images. The model offered low segmentation performances while the guidewire class was invisible to the model.

The research gaps identified motivate the development of a DL-based technique for simultaneous multi-class segmentation in cardiovascular angiograms. Built on the existing studies [9, 11], a dual-part multi-lateral branched network (MLBNet) is developed on previous MLBNet designs, and validated for segmentation of tool, vessel, and background structures in catheterisation angiogram scenes. A distinct step in this study is equipping the encoder and decoder sub-networks with multiple lateral branches for shared representation learning. The main contributions of this work are in threefold. **First**, a dual-part MLBNet with encoder–decoder lateral branching is proposed for simultaneous multi-class pixel-level segmentation. This is aimed to address limitations such as sequential multi-head branching and direct shared decoder usage of earlier works. **Second**, a hybrid objective function with structure-specific loss terms in the decoder to prevent vulnerability to class dominance and imbalance issues in cardiovascular catheterization angiogram datasets. **Third**, comprehensive evaluation studies performed and reported to evaluate the dual-part MLBNet on multiple catheterization angiogram datasets in different anatomies.

The remainder of this paper is organized as follows: the dual-part MLB-Net design is introduced in Section II, while the model implementation and training processes are presented in Section III, and experimental studies carried out for model evaluation are reported in Section IV. Lastly, conclusion of the study and future works are in Section V.

## 2. Multi-Class Segmentation Technique

Multi-class segmentation of angiogram background and foreground structures is central to procedural analytics in cardiovascular theranostics. The multi-class segmentation problem formulation and the MLBNet design proposed for pixel-level segmentation of structures in angiograms is presented in this section.

### *2.1. Problem Formulation*

Multi-class segmentation in catheterization angiograms is challenging since the scene structures are degraded by noise artifacts, low contrast, structural overlap, and severe class imbalance. Let $x \in \mathbb{R}^{H \times W \times C}$ denote an angiogram with dimensions of $H$ and $W$, and $C$ is the number of input channels. The goal is to learn a mapping $f_\Theta: x \to \hat{Y}$ where $\hat{Y} \in [0,1]^{H \times W \times K}$ is a dense class-probability map over $K$ procedure-specific scene structures with $K$ classes, and $Y \in \{0,1\}^{H \times W \times K}$ is the corresponding pixel-level one-hot encoding ground-truth label. It is herein hypothesised that laterally-branched encoding and decoding neural blocks could support multi-class segmentation with improved and stable performances. This would enhance task-specific decoder learning for multi-class segmentation.

### *2.2. Dual-Part MLBNet Design*

The dual-part MLBNet design is presented in Fig. 1, with the proposed encoder–decoder multi-lateral branching network design explained.

#### *2.2.1 Encoder Sub-Network*

Lateral branching in the encoder involves using a multi-block sub-network built on an Xception backbone [11].

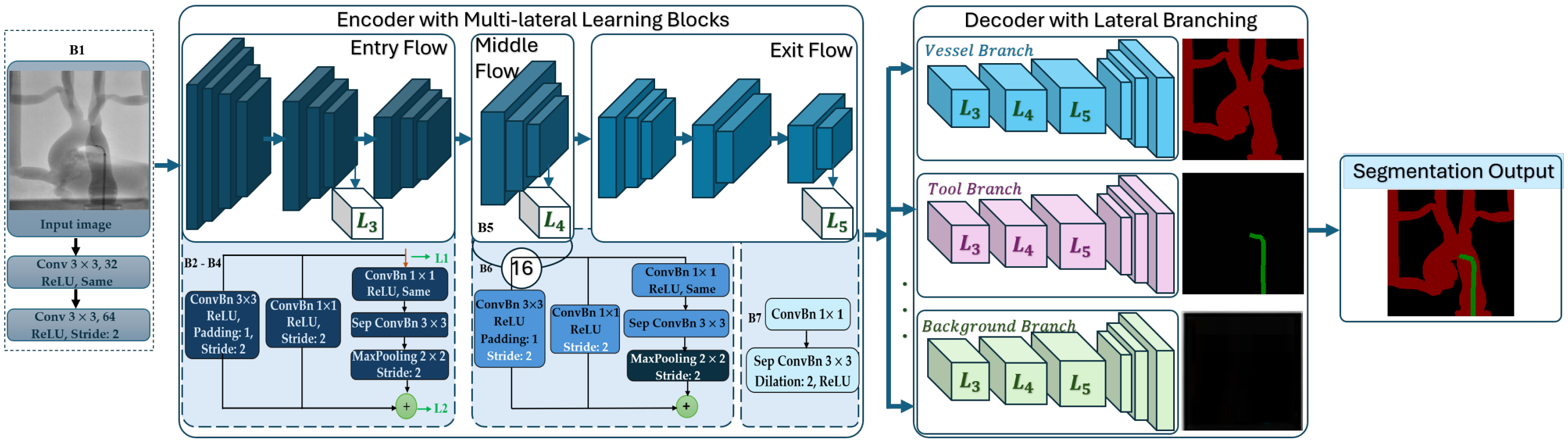

Figure 1: Proposed Dual-Part Multi-Lateral Branched Network for Multi-Class Endovascular Scene Segmentation

The input angiograms are initially processed through a shallow stem block, denoted $B_1$ in Fig. 1, and the features obtained stem block, denoted $B_1$ in Fig. 1, and the features obtained are passed through consecutive laterally branched segments in the encoder. The stem block executes a serial function: $F_1 = \phi_{\text{stem}}(x)$ , where $\phi_{\text{stem}}$ is composed of convolution, batch normalization and ReLU activation, to obtain low-level spatial features from an input angiogram scene.

## A. *Lateral Encoder Branching*

Each lateral branch in the encoder is defined with three parallel convolutional paths to obtain diverse feature maps. The parallel branches are used for complementary feature learning and representations based on input feature maps from the images received. Depthwise convolutions are applied to preserve local edge and line responses crucial for thin elongated structures. The parallel branches have unique outputs ($U_\ell^{(i)}$) defined in Eq. 1, and each branch is followed by batch normalization and ReLU activation. However, the middle branch is extended with separable convolution to improve the feature diversity, while max-pooling operation is also performed to reduce feature dimensionalities. Where $F_\ell$ is an input feature map to the encoder block $\ell$, $\phi_\ell^{(i=1)}$is a $1 \times 1$ projection branch, $\phi_\ell^{(i=2)}$ is the main separable-convolution branch, and $\phi_\ell^{(i=3)}$ is either a $3 \times 3$ convolution or an identity branch depending on the block.

$$U_\ell^{(i)} = \phi_\ell^{(i)}(F_\ell) \quad (1)$$

A fused output $\left(\tilde{F}_\ell = \mathcal{F}_\ell\left(U_\ell^{(1)}, U_\ell^{(2)}, U_\ell^{(3)}\right)\right)$ is obtained through channel concatenation followed by projection or additive residual fusion, when dimensions align, and the residual update (Eq. 2) is performed, where $\psi_\ell(\cdot)$ is an identity or projection mapping used to match dimensions. In addition, atrous separable convolutions with adaptive dilation rates are used to obtain contextual multi-scale features and reduce network parameters [11].

$$F_{\ell+1} = \tilde{F}_\ell + \psi_\ell(F_\ell) \quad (2)$$

## B. *Multi-flow Structure*

The encoder blocks are organized into three connected fragments organized as *entry flow*, *middle flow*, and *exit flow*. The entry flow corresponds to blocks $B_2$–$B_4$ and it is used to capture low-level and intermediate contextual cues for descriptive feature representations. Dilation and field-of-view (FoV) expansion are applied to maintain consistent output dimensions across branches. The receptive-field growth per block is obtained with a scaling factor ($\alpha$). For progressive context aggregation, $\alpha > 1$ is used. The middle flow, specifically block $B_5$, encodes class-specific features using operations similar to $B_6$, excluding max-pooling and stride, which are repeated 16 times to learn discriminative features and improve feature-map diversity during training. The exit flow, corresponding to $B_6$–$B_7$, is applied to refine feature representation using pointwise and atrous convolutions. The residual features from $B_3$, $B_5$, and $B_7$ are aggregated as $L_3$, $L_4$, and $L_5$, and bilinearly upsampled (in Eq. 3, where $s \in \{F_{B3}, F_{B5}, F_{B7}\}$ denote the skip features, and $\mathcal{U}_s(\cdot)$ denotes bilinear upsampling and channel alignment.) to support multi-scale deep supervision in the decoder. This multi-scale fused feature map is then passed to the decoder sub-network.

$$F_{\text{agg}} = \sum_{s \in \{B3, B5, B7\}} \mathcal{U}_s(F_s), \quad (3)$$

### 2.2.2 *Decoder Sub-network*

In cardiovascular catheterization, tool pixels are usually contained within or along a vessel structure, while vessel forms a continuous tree-like structure in the catheterization scenes. Hence, multi-class structure pixels overlap, and the use of single shared segmentation head may not provide learning stability for such class interactions. Rather than relying on a single shared softmax head or sequential heads [9, 11, 19] in prior studies, the decoder sub-network in this study extends with class-specific lateral decoder branching. This design is motivated since target structure classes are geometrically coupled.

For context, the class-specific lateral branching works by having a multi-head architecture where one decoder branch is oriented towards tool class, another towards vessel class, and the last branch for scene background region external to foreground classes. This lateral decoder design preserves class-aware structure segmentation and prevents redundant heads in the earlier studies [9, 11, 19]. To learn meaningful scene representation, the class-specific head to each decoder outputs $Z^{(c)} = \phi_d(F_{\text{agg}})$ for class $(c)$, where $\phi_d$ is a scene-wide decoder for the $K$-class segmentation task. The final segmentation output is obtained as $\hat{Y} = Softmax(L)$, using a late fusion of class logits ($L \in \mathbb{R}^{H\times W\times 3}$) from the decoders assembled into a single logit tensor given in Eq. 4. $W_f^{(\cdot)}$are fusion weights and $\Pi_b, \Pi_v, \Pi_t$ are projection operators that map the branch-specific outputs for background, vessel and tool pixels into a global class space.

$$L = Concat\left(W_f^{(b)}\Pi_b(Z^{(b)}), W_f^{(b)}\Pi_b(Z^{(b)}), W_f^{(b)}\Pi_b(Z^{(b)})\right) \quad (4)$$

This strategy allows the network to combine global scene context with class-specific structure refinements while also preserving the multi-lateral branching principles.

### 2.2.3 *Class-Aware Scene Objective Function*

As standard cross-entropy function could be vulnerable to background dominance and class imbalance, we follow the idea of using hybrid loss function and propose a class-aware scene (CAS) objective function. Let the three lateral decoder branches produce class-specific probability maps $P^{(b)}$, $P^{(v)}, P^{(t)} \in [0,1]^{H\times W}$ for background, vessel and tool pixels in an angiogram scene, respectively. If the complete multiclass one-hot ground truth is denoted by $Y$, the fused main decoder can be said to output a multi-class scene prediction $\hat{Y} \in [0,1]^{H\times W\times 3}$ where $Y^{(b)}$, $Y^{(v)}$, $Y^{(t)}$ are the corresponding one-vs-rest binary targets. To preserve class discrimination, the CAS function transforms the MLBNet approach to a structured class-aware segmentation network.

In addition to categorical cross-entropy, multi-class focal loss was optimized with connectivity-sensitive supervision. In angiographic scene analysis, endovascular tools, vessels, and background are both geometrically coupled rather than independent semantic classes. To enable class-specific learning in the MLBNet, CAS operates loss components for tool, vessel, and background classes in an angiogram scene. For branch $c \in \{t, v, b\}$, structured focal cross-entropy loss, defined in Eq. 5, is used to optimize each branch as a one-vs-rest classifier with structured non-target supervision, thus all pixels constrain the branch, while non-target classes are adaptively weighted. This retains the discriminative strength of cross-entropy and reweighs the low frequency target classes against competing foreground and dominant background classes. Then a branch-wise Tversky loss, given in Eq. (6), is used improve overlap under imbalance.

$$\mathcal{L}_{\text{SFCE}}^{(c)} = -\frac{1}{N}\sum_i \left[\alpha_c Y_i^{(c)}(1-P_i^{(c)})^\gamma \log P_i^{(c)} + \sum_{k\neq c}\beta_{c,k}\, Y_i^{(k)}(P_i^{(c)})^\gamma \log(1-P_i^{(c)})\right] \quad (5)$$

$$\mathcal{L}_{\text{Tv}}^{(c)} = 1 - \frac{TP_c + \epsilon}{TP_c + \delta_c FN_c + (1-\delta_c)FP_c + \epsilon} \quad (6)$$

where $\alpha_c$controls the positive-class importance, and $\beta_{c,k}$ controls the penalty from each non-target class $k$. Also, $\gamma$ is a focal focusing factor, and $\delta_c$ controls the penalty trade-off between the false negatives and false positives for class $c$.

### A. *Tool Branch Loss Term*

As the decoder design focuses on task-specific learning, one of the decoder branches is optimized to supervise tool related terms using tool Dice and skeleton-consistency functions. The tool branch learns intravascular plausibility of guidewire or catheter in an angiogram scene using Eq. 7. It defines skeleton-consistency terms $\left(\mathcal{L}_{\text{skel}}^{(t)}\right)$ in Eq. 8 that accounts for structure thinness, connectivity and centerline, and structural containment $\left(\mathcal{L}_{\text{in}}^{(t)}\right)$ in Eq. 9. The latter helps to account for catheterization scene parsing knowledge that tool prediction must lie within the vessel support. Thus, tool predictions outside the vessel support can be penalized.

$$\mathcal{L}_t = \mathcal{L}_{\text{SFCE}}^{(t)} + \lambda_{t1}\mathcal{L}_{\text{Tv}}^{(t)} + \lambda_{t2}\mathcal{L}_{\text{skel}}^{(t)} + \lambda_{t3}\mathcal{L}_{\text{in}}^{(t)} \quad (7)$$

$$\mathcal{L}_{\text{skel}}^{(t)} = 1 - \frac{2\sum_i \text{Skel}(P_i^{(t)})\,\text{Skel}(Y_i^{(t)}) + \epsilon}{\sum_i \text{Skel}(P_i^{(t)}) + \sum_i \text{Skel}(Y_i^{(t)}) + \epsilon} \quad (8)$$

$$\mathcal{L}_{\text{in}}^{(t)} = \frac{1}{N}\sum_i \max\left(0, P_i^{(v)} - Y_i^{(t)}\right) \quad (9)$$

For non-equivalent supervision, background is treated as stronger negative than vessel in this branch, i.e. $\beta_{t,b} > \beta_{t,v}$, because vessel pixels are anatomically related to the tool and should not be penalized in the same way as background.

### B. *Vessel Branch Loss Term*

Similarly, a combined Dice-IoU loss term is applied with boundary-aware constraint on another decoder branch to understand vessel structures. The vessel branch is designed to emphasize anatomical region coherence, tool boundary and support continuity using the loss terms in Eq. 10. $\mathcal{L}_{\text{bd}}^{v}$ defined in Eq. 11, is a boundary-support term set to bound tool pixels within the vessel pixels, and $\nabla$ denotes a spatial gradient operator. Background is the strongest negative, while tool is a structured internal competitor: $\beta_{v,b} > \beta_{v,g}$.

$$\mathcal{L}_v = \mathcal{L}_{\text{SFCE}}^{(v)} + \lambda_{v1}\mathcal{L}_{\text{Tv}}^{(v)} + \lambda_{v2}\mathcal{L}_{\text{bd}}^{(v)} \quad (10)$$

$$\mathcal{L}_{\text{bd}}^{v} = \frac{1}{N}\sum_i \| \nabla P_i^{(v)} - \nabla Y_i^{(v)} \|_1 \quad (11)$$

*C. Background Branch Loss Term*

The background branch is designed to be discriminative without overwhelming the other branches, as given in Eq. 12. To reduce background dominance, its positive weight is intentionally moderated: $\alpha_b < \alpha_v, \alpha_t$. Similarly, vessel is penalized more strongly than tool in the background branch since the vessel would occupy larger contiguous foreground support: $\beta_{v,b} > \beta_{t,b}$. This design ensures that each branch is fully constrained by all pixel classes, but target and non-target classes are not treated equivalently.

$$\mathcal{L}_b = \mathcal{L}_{\text{SFCE}}^{(b)} + \lambda_{b1}\mathcal{L}_{\text{Tv}}^{(b)} \quad (12)$$

*D. Fused Multiclass Scene Supervision*

For the fused scene head should remain the dominant segmentation objective, a multiclass scene supervision is used as a unified learning strategy where the scene branch is trained against a multi-class segmentation label map that jointly represents background, vessel, and tool classes. This allows the model to learn global scene context and inter-class relationships under one fused semantic objective. The unified supervision is optimized with class-balanced cross-entropy ($\mathcal{L}_{\text{CBCE}}$) and Dice functions for multi-class supervision, as given in Eq. 14. The right-most term is used to ensure containment prior to preserve strong multi-class discrimination where standard categorical cross-entropy is vulnerable.

$$\mathcal{L}_m = \mathcal{L}_{\text{CBCE}}(Y, \hat{Y}) + \lambda_{m1}\mathcal{L}_{\text{Dice}}^{\text{mc}}(Y, \hat{Y}) + \lambda_{m2}\frac{1}{N}\sum_i \max(0, \hat{Y}_{i,g} - \hat{Y}_{i,v}) \quad (13)$$

*E. Total Loss*

The overall training loss function is therefore a weighted sum of the loss terms, as given in Eq. 14, where $\lambda_{\cdot}$ are scalar hyperparameters. This function involves adaptive task balancing to reduce gradient interference across the four objectives $\mathcal{L}_t, \mathcal{L}_v, \mathcal{L}_b, \mathcal{L}_m$ using learnable uncertainty-based weights. $\sigma_q$ is a scalar that allows the network to rebalance branch losses during training for task $q$. The optimal hyperparameter values can be determined through a grid search or ablation studies.

$$\mathcal{L}_{\text{total}} = \sum_{q\in\{t,v,b,m\}} \left(\frac{1}{2\sigma_q^2}\mathcal{L}_q + \log\sigma_q\right) \quad (14)$$

## 3. Network Implementation and Training

### *3.1. Network Implementation*

The proposed dual-part MLBNet design is implemented with Tensorflow® Keras® and validated on catheterization datasets. The multi-lateral learning concept is applied to both the encoder and decoder sub-modules, as described in Section II. The encoder part was implemented as a modified Xception-style backbone with the repeated lateral branches fused for complementary representation learning. For decoder part, class-specific learning is approached through shared upper decoder trunk for class-aware segmentation. In a three-branch decoder design, the first branch could be used for tool segmentation, the second branch for vessel segmentation, and the third branch for the background. All branch heads receive shared feature maps from the encoder blocks. The latter branch heads are built on earlier decoder layers in addition to convolution and dropout layers. Thus, different feature maps are received by the three decoders, and this helps to prevent class-specific model overfitting.

### *3.2. Network Training Protocol*

The proposed MLBNet model was trained for pixel-level segmentation in three catheterization datasets. Each dataset was randomly partitioned for model training and testing. The images were resized to 256 × 256 pixels for network input. Adam optimizer with parameters: $\beta_1$ = 0.9, $\beta_2$ = 0.999, and $\varepsilon = 10^{-6}$ were applied during training. Based on preliminary studies, an initial learning rate of $10^{-4}$ was used as it produced stable validation behaviour across the datasets. The learning rate dynamically adjusted using drop and decay factors of 1 and 0.95, respectively, to improve training convergence. Gradient clipping with *clipnorm* = 1.0 was explored for improved optimization. During model fitting, training and validation data splits were generated at 80% to 20%, respectively. These were processed through a batch-based angiogram loader with one-hot label encoding applied for all classes. The CAS loss terms designed in Section III was applied to address limitations of standard categorical cross-entropy under strong class imbalance and background dominance. Thus, branch-specific losses were assigned to the tool, vessel, and background branch heads.

The network was trained multiple times with a batch size of 16 and 50 epochs on High-Performance Computing facility with NVIDIA A100 GPU cards and 48 GB memory. The training process involved active callbacks with checkpoint set on validation-loss for learning-rate reduction and early stopping. Optimally tuned multi-head weights: 0.15, 0.2, 1.0 were used for the three decoder heads, respectively. This reduced gradient interference and yielded the best results. While the weights only favoured **Dataset 1-2**, weights: 0.75, 1.25, 1.0 stabilized the performance in **Dataset 3**. Data preprocessing for model training follows the same setting data splits and input resolution. Guided by F1-score, and mean Intersection over Union (mIoU) metrics, model giving the best segmentation quality was selected for experimental and ablation studies. The best performing models obtained after training on each dataset were evaluated with the corresponding test sets.

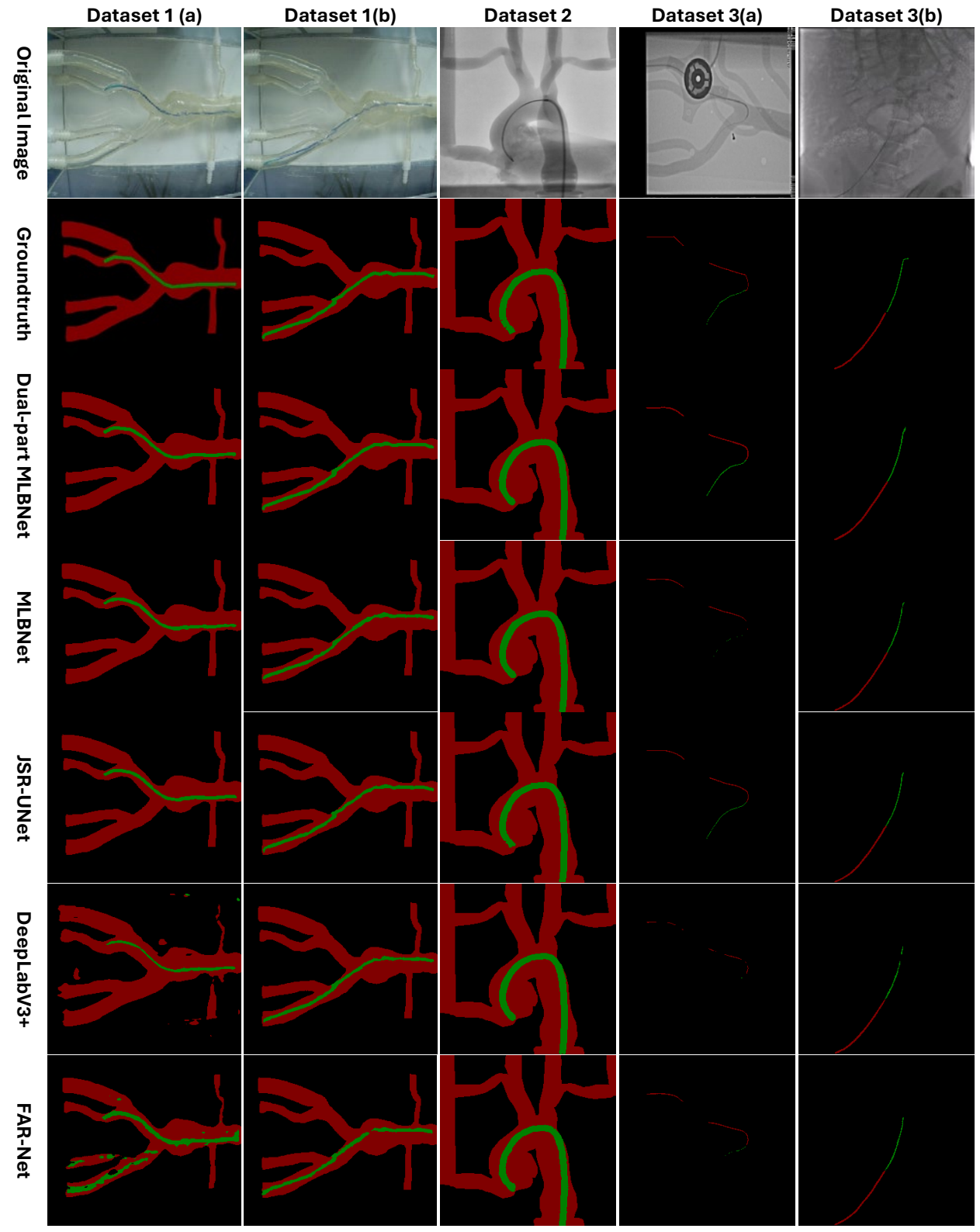

Figure 2: Segmentation results for selected frames

### *3.3. Datasets and Preprocessing*

The model was validated with three angiogram datasets obtained during different catheterization procedures. These includes both *in-vitro* and *in-vivo* datasets described herein. The *in-vitro* datasets consist of proprietary camera-based 857 and 874 images of catheterization scenes saved during manual and robot-assisted tool navigation (**Dataset** 1a-b) in a silicone phantom, respectively. Each image frame has 640 × 480 pixels, and multi-class annotation was manually done with LabelMe 6.3.1. A public dataset with frames extracted from fluoroscopy video (**Dataset 2**) was also used in this study. The dataset consists of 2,000 angiograms, each with 256 × 256 pixels, recorded in silicone-based aortic phantom [13]. Annotation for this dataset only includes tool masks in its original form; thus, extra step was taken to add vessel masks to the catheterization scenes. This was achieved with custom code written to propagate a self-created vessel mask done for first frame to the later frames. Lastly, CathAction dataset, largest public fluoroscopic dataset with angiograms recorded during catheterization in silicone-based phantom (**Dataset 3a**) and animal experiments (**Dataset 3b**) was also used in this study [20]. The animal setup was made identical to human procedures, surgeons completed catheterisation in the left subclavian, left common carotid, and right common carotid vessels with commercial catheter and guidewire. In this dataset, tool pixels were manually annotated, thus could support multi-class foreground segmentation tasks only.

TABLE I. MODELS' PERFORMANCES ACROSS THE DATASETS (%)

| Test set | Acc (%) | F1-S (%) | mIOU (%) for Case | | | | |
|---|---|---|---|---|---|---|---|
| | | | #1 | #2 | #3 | #4 | #5 |
| Data 1a | 99.30 | 96.94 | **94.21** | 91.15 | 89.78 | 94.21 | 39.26 |
| Data 1b | 99.27 | 96.84 | **94.03** | 89.55 | 89.83 | 94.03 | 86.69 |
| Data 2 | 99.20 | 97.28 | 94.86 | **94.89** | 93.64 | 80.66 | 23.86 |
| Data 3a | 99.51 | 66.32 | **54.12** | 49.76 | 50.40 | 17.32 | 8.56 |
| Data 3b | 99.30 | 77.29 | **66.37** | 61.81 | 62.53 | 16.48 | 17.47 |

The masks for first two datasets (1a, 1b, and 2) include background–vessel–tool pixels, thus multi-class annotation is achieved as background masks encoded by "0", vessel masks by “1”, and tool masks by “2”. For the CathAction dataset that rather has background–catheter–guidewire masks, background masks were encoded by "0", catheter masks by “1” and guidewire masks by “2”. The three datasets were pre-processed, used to train and validate the dual-part MLBNet model separately. In addition, data-agnostic tests were done by evaluating the performance of a model trained with one data on others without re-training.

## 4. Experimental Studies and Evaluations

The dataset details and the segmentation results obtained with the dual-part MLBNet are presented.

### *4.1. Model Evaluation and Analysis*

To assess the model’s segmentation quality, both full-scene and foreground-sensitive metrics were selected to evaluate the model. These include accuracy, F-score, and mIoU. Background-excluded versions of the metrics were considered as overall performance could misinform due to the background pixel dominance in catheterization scenes. Fig. 2 shows the segmentation outputs from the dual-part MLBNet for a random frame in the test set of each dataset. As shown in third row of Fig. 2, the model could accurately delineate and track three-class structures in all the datasets, achieving high overall classification performance across all the test sets with higher macro F1 for **Dataset 1** (96.94%), **Dataset 2** (97.28%) and lower macro F1 (66.32%, 77.29%) for CathAction phantom and animal test sets in **Datasets 3.**

Across all test sets, the models got a strong segmentation experience for the background class, with F1-score range of 99.67–99.90%. This shows a huge class imbalance skewing towards background, and bias to interpret the model using the accuracy values in Table I. However, the other metrics show the dual-part MLBNet architecture could also localize the foreground structures. The confusion matrices in Fig. 3 explains the model had a lesser foreground performance for background–catheter–guidewire segmentation compared to the background–vessel–tool task.

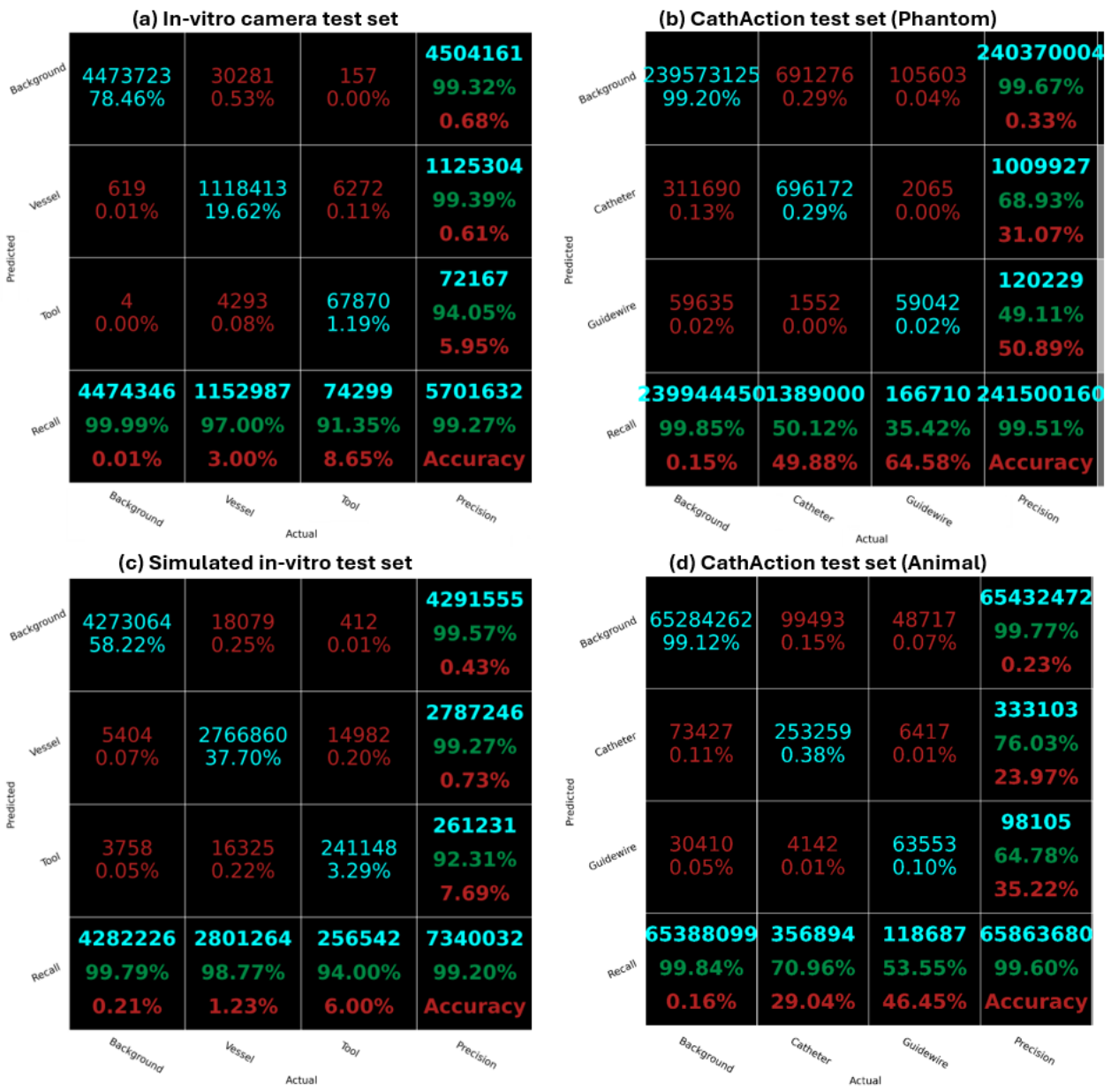

Figure 3: Performance analysis of models for four datasets

The foreground structure analyses show high error rates, as explained with precision (50.89%, 35.22%) and recall (64.58%, 46.45%) for **Dataset 3**, with *Phantom* and *Animal* angiograms, respectively. The results show the model struggles to correctly classify catheter and guidewire, tools with similar appearances, and confuse them with one another or even the background.

### *4.2. Analyses of Models' Domain Adaptation*

The dual-part MLBNet architecture's ability for domain adaptation was evaluated for models built on the different datasets. For this experiment, the model built on training of dataset was applied for multi-class segmentation on the test set of others. Cross-domain performances were analyzed using tool micro-average precision-recall curves (AUPRC) in results in Fig. 4. The curves show model transferability between visually similar domains, for instance the camera-based phantom **Datasets 1,** tool AUPRC = 0.863. However, the transferability degraded in cross-domain heterogeneous evaluation such as from animal and phantom sets, or vice-versa, in **Dataset 3**. Thus, dual-part MLBNet generalises well across visually similar domains but poorly in the presence of domain shifts caused by differences in vessel visibility, tool contrast, anatomical structure, and imaging.

### *4.3. Ablation Studies*

Studies were performed to analyze the decoders' multi-head design and the learning functions.

#### *A. Multi-head and Branch-Specific Loss Contributions*

The default multi-head setup and the branch specific loss terms in the Eq. 14 were used was taken as *Case #1* with the mIOU values presented in Table I. *Case #2* is a follow-up ablation conducted to determine how the branch specific objective function contributes to the models' segmentation performances. For this, the CAS objective function (in Eq. 14) was substituted with standard categorical cross-entropy. The results obtained showed that the branch specific three-decoder setup achieved better and more stable segmentation performances for four datasets; its performance for **Dataset 2** was closer to when the standard categorical cross-entropy was used. The proposed branch specific objective function has the best and most stable segmentation performance.

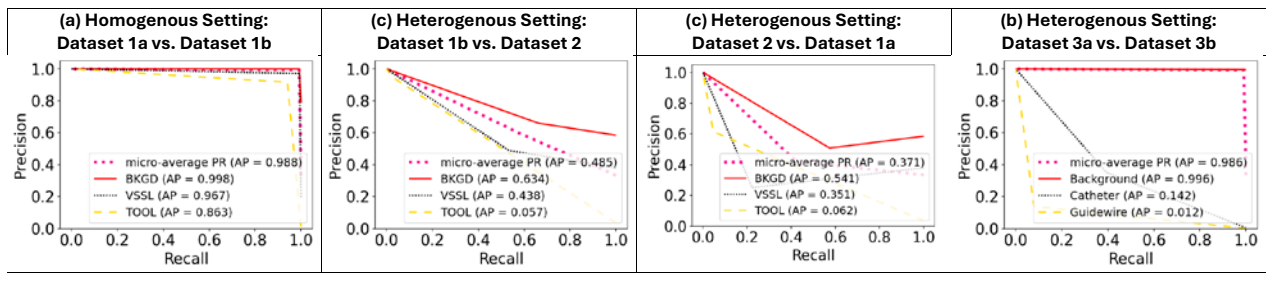

Figure 4: Segmentation results for selected frames in camera data

We also checked if the models' improved performance is majorly from the multi-decoder architecture (Case #3). The multi-head architecture was set to use one decoder and the standard categorical cross-entropy. Compared with the outcomes in *Cases #1* and *Case #2*, we found that a single-decoder configuration variant has a less stable foreground segmentation performance. While a typical one-decoder configuration achieved high performance on **Datasets 1-2**, (*i.e.*, the simpler silicone-based data), a clear degradation is observed in the foreground segmentation performances on **Dataset 3**, the challenging CathAction data. These indicate that the proposed multi-head architecture with the branch specific loss terms contributes to class-specific learning.

#### *B. Analysis of Multi-Head Segmentation Outputs*

While segmentation with multi-head branching could be more robust than single decoder architecture, differences in decoder perturbations may cause variations in segmentation performances of the heads. To verify the top performing head, the vessel-head and tool-head branches for pixel-level segmentation, as given in *Case #4* and *Case #5* of Table I, respectively. Note that the legacy vessel-head and tool-head branches were respectively designed to segment catheter and guidewire pixels in the CathAction dataset. We found the pixel-level predictions from the scene-head branch gave the best performances across all the datasets (*Case #1*) followed by the tool-head branch, while the vessel-tool gave the least pixel-level prediction performances.

### *4.4. Performances for Different DL Architectures*

Existing deep learning architectures used for binary tool segmentation in previous study [11] were evaluated and their performances compared with that of the proposed dual-part MLBNet. The segmentation performances that were obtained for these models are presented in Table II.

TABLE II. COMPARISON OF DL ARCHITECTURES (METRICS IN %)

| Architecture | Dataset 1a | | Dataset 2 | | Dataset 3b | |
|---|---|---|---|---|---|---|
| | mIoU | Tool Error | mIoU | Tool Error | mIoU | Tool Error |
| **Proposed** | **94.21** | **5.80** | **94.86** | **7.69** | **66.37** | **24.08** |
| MLBNet | 89.78 | 14.81 | 93.64 | 10.94 | 62.53 | 33.29 |
| JSR-UNet | 92.76 | 10.30 | 78.93 | 11.29 | 63.60 | 31.77 |
| FAR-Net | 91.90 | 10.91 | 90.40 | 12.83 | 60.71 | 34.66 |
| DeepLabV3+ | 80.68 | 23.23 | 93.27 | 11.05 | 62.25 | 33.25 |

First, the models' segmentation outputs displayed in Fig. 2 (Rows 3 – 7) show dual-part-MLBS-Net models are robust for scene-level angiogram segmentation. As presented in Table II, the proposed architecture has best metrics in all architectures compared with. Tool (precision) error is the proportion of false tool prediction (catheter and guidewire are combined in **Dataset 3**) divided by the total predicted tool pixels in a test set. The dual-part MLBS-Net has the lowest error values, followed by JSR-UNet and the earlier MLBNet models. This informs the models' ability for tool detection and in separating the foreground structures with respect to non-tool classes.

## 5. Conclusion and Future Works

In this study, a dual-part MLBNet architecture was proposed and validated for multi-structure segmentation in catheterization scenes. Models supporting different class-aware segmentation were evaluated on different datasets to aid intuitive scene parsing in cardiovascular catheterisation procedures. Compared with earlier methods, the proposed architecture integrates shared encoder with partially shared upper decoder trunk, and branch-specific decoder blocks. Evaluation results from both *in-vitro* and *in-vivo* datasets showed stable segmentation performances, and feasibility for cross-domain segmentation of multiple structure in the different catheterisation datasets were also demonstrated.

The architecture combines shared feature extraction with differentiated representations observed for the three heads. A high linear centered kernel alignment (CKA = 1.00) was observed by analysing the shared auxiliary branch features, whereas cross-representation with the main branch showed a lower CKA, indicating differentiation at the analyzed downstream feature locations. In addition, the limited transparency of deep learning systems remains a key concern. The dual-part MLBNet achieved promising results across the datasets. Currently, the multiple decoder head weights are tuned separately for each dataset in this study. A learnable approach could enable data-agnostic modeling. Lastly, there is need to consider scenes where other devices such as stent and ballon used during the clinical practice.

**Acknowledgment**: AI-assistance was used to support model implementation, code debugging and drafting. These were reviewed, modified, and validated by the authors.